\RequirePackage{fix-cm}
\documentclass[twocolumn]{svjour3}          % twocolumn

\smartqed  % flush right qed marks, e.g. at end of proof

\usepackage{graphicx}
\usepackage{amssymb}
\usepackage{natbib}
\usepackage{algorithm}
\usepackage{algorithmic}
\journalname{IJCV}

\begin{document} \sloppy
\title{A General Preprocessing Method for Improved Performance of Epipolar Geometry Estimation Algorithms}
\author{Maria Kushnir \and Ilan Shimshoni}
\institute{
		Maria Kushnir
			\at Department of Information Systems, University of Haifa, 31905 Haifa, Israel.\\
			\email{mkushn01@campus.haifa.ac.il}
		\and
		Ilan Shimshoni
			\at Department of Information Systems, University of Haifa, 31905 Haifa, Israel.\\
			\email{ishimshoni@mis.haifa.ac.il}
}

\date{Received: date / Accepted: date}

\maketitle

\begin{abstract}
In this paper a deterministic preprocessing algorithm is presented, whose output can be given as input to most state-of-the-art epipolar geometry estimation algorithms, improving their results considerably. They are now able to succeed on hard cases for which they failed before. The algorithm consists of three steps, whose scope changes from local to global. In the local step it extracts from a pair of images local features (e.g. SIFT). Similar features from each image are clustered and the clusters are matched yielding a large number of putative matches. In the second step pairs of spatially close features (called 2keypoints) are matched and ranked by a classifier. The 2keypoint matches with the highest ranks are selected. In the global step, from each two 2keypoint matches a fundamental matrix is computed. As quite a few of the matrices are generated from correct matches they are used to rank the putative matches found in the first step. For each match the number of fundamental matrices, for which it approximately satisfies the epipolar constraint, is calculated. This set of matches is combined with the putative matches generated by standard methods and their probabilities to be correct are estimated by a classifier. These are then given as input to state-of-the-art epipolar geometry estimation algorithms such as BEEM, BLOGS and USAC yielding much better results than the original algorithms. This was shown in extensive testing performed on almost 900 image pairs from six publicly available datasets.
\end{abstract}

\section{Introduction} \label{intro}
Epipolar geometry estimation from image pairs with partial scene overlap is a basic problem in computer vision. It is used as a component of many important applications such as vision based robot navigation, structure from motion (SfM) and other multiple view geometry applications.

This problem has attracted considerable interest in the computer vision community, interest which continues till this day. Most of the successful algorithms are based on an initial step, in which local features are detected in both images. For each detected feature a local descriptor is  computed. These features are then matched based on their local descriptors. For each putative match a prior probability or score is estimated. These putative matches and scores are given as input to the algorithm~\citep{LO_RANSAC,PROSAC,BLOGS,BEEM,USAC,GuidedRansac}. Even though successful algorithms have been proposed to address this problem, it still remains an active field of research. This is because there are several reasons why input given to these algorithms may be challenging. As pointed out for example by~\citet{SIFT} and tested extensively by~\citet{mikolajczyk2005comparison}, as the angle between the viewing directions increases, the appearance of local descriptors changes, making them hard to match. Thus, wide baseline images are hard inputs for the algorithms. Urban scenes are also challenging for such algorithms. In such scenes features such as for example windows are repeated several times. In such cases it is hard for the local matching algorithm to match the window in the first image to its corresponding window in the second image. In both these types of cases the percentage of correct matches (inliers) from the  set of putative matches is low. When the probabilities are taken into account, the problem is that  the percentage of correct matches with high prior probabilities is low. In these cases, even state-of-the-art algorithms tend to fail.

For that reason, in this paper, instead of trying to propose a new epipolar geometry estimation algorithm, we present a preprocessing step which is given as input two images and returns a set of putative matches with their associated probabilities. Our method was extensively tested on almost 900 image pairs from different datasets: ZuBuD dataset~\citep{ZuBuD}, BLOGS dataset~\citep{BLOGS_DB}, USAC dataset~\citep{USAC_DB} and Open1, Open2 and Urban datasets~\citep{SOREPP_DB}. Our results are much better than those obtained by the standard initial steps of state-of-the-art algorithms.  Consequently, when our output is given as input to them (BEEM~\citep{BEEM}, BLOGS~\citep{BLOGS} and USAC~\citep{USAC} in this paper) they outperform the same algorithms operating on their regular input. Our output is general and can be incorporated within many other algorithms such as~\citet{LO_RANSAC,PROSAC,GuidedRansac}.

The algorithm starts with standard techniques of detecting local features and extracting putative correspondences from them. Using this input we propose a new concept consisting of three steps, running from local to global. In the local step features are clustered together in each image. Clusters with similar features from the first image are matched to clusters of features from the second image and vice versa. The result of this step is a large set of putative matches, most of which are incorrect. In the second step we match pairs of spatially close features (2keypoints) in the first image, to corresponding pairs of features (found in the first step) from the second image. For each 2keypoint match a short descriptor is generated, characterizing the quality of the match. Using a classifier we trained on data from several image pairs, for each 2keypoint match the  probability of being correct is estimated. The highest $K_{2kp}$ 2keypoint matches are chosen. Here already the percentage of correct 2keypoint matches is much higher then was recovered in the first step. In the global step, the 2keypoint matches are used to generate a large number of possible fundamental matrices. For each putative match from the first step we calculate the number of fundamental matrices it supports. Finally we combine putative matches generated by standard methods with those found by our method, and estimate their probabilities to be correct, using a simple classifier.

The correlation between these probabilities and the ground truth inlier-outlier labels is much higher. As a result, when we submit the putative matches and the computed probabilities as input to algorithms from the guided RANSAC family, much better results are obtained on challenging datasets. For example, the performance of all three algorithms when run on the Open2 dataset~\citep{SOREPP_DB} increased considerably. The number of image pairs they succeeded on increased by between 62\% and 239\% relative to the original performance of those algorithms. This demonstrates the fact that our algorithm improves the quality of the input significantly, resulting in better results of the basic algorithm. Similar improvement was obtained when the algorithm was run as a preprocessing step of USAC on the Urban dataset~\citep{SOREPP_DB}.

The paper continues as follows. In Section~\ref{sect:related}, we review related work concentrating mainly on how the quality of the input affects the performance of the algorithm. Section~\ref{sect:outline} presents the overview of our method, while the details are given in the next section. Experimental results are presented in Section~\ref{sect:experiments}. We conclude in Section~\ref{sect:conc}.

\section{Related work}\label{sect:related}

In reviewing related work we will concentrate on how the quality of the input effects the algorithm's performance and not on the various components of the algorithms.

We will first consider PROSAC~\citep{PROSAC} and USAC~\citep{USAC}. The algorithm is given as input a set of putative matches ordered by a score or prior probability. Under this general framework the set of putative matches can be ordered for example using the distance ratio $d_r$ method introduced by~\citet{SIFT}. The models are generated in an order consistent with the order  of the matches used to generate them. Once a model is generated it is verified using the Statistical Probability Ratio Test (SPRT). The putative matches are tested until the SPRT reaches a decision on whether the model is correct or not. Thus, when the beginning of the list (matches with high scores) is contaminated by a large number of outliers, the number of required iterations increases considerably. When the number of iterations of the algorithm is limited, this also increases the probability of failure. On the other hand, a list consisting of a large number of matches does not effect the running time, since the SPRT process usually reaches a decision quite early in the verification procedure.

In algorithms from the Guided RANSAC family~\citep{GuidedRansac,BEEM} the subset of matches used for model generation is chosen according to their probability. Thus, the performance of the algorithm is similar to that of PROSAC. When the list contains a large number of outliers with high probabilities, the chances of the algorithm to fail are high.

A similar behavior occurs in BLOGS~\citep{BLOGS}. There also, in the global search step, a model is computed from a minimal subset of matches according to a score. Thus, the probability of finding a model consisting only of inliers depends on the quality of the prior scores. Specifically in BLOGS, a new method for putative match ranking was introduced, and their scores are referred to as similarity weights $\left\{t_k\right\}$.

Thus, for all these algorithms, if we can assign more accurate probabilities to the putative matches, the algorithms performance should improve considerably. This is exactly the goal of the algorithm we suggest here.

We would also like to review two other algorithms which address the problem of matching images containing scenes with repeated structures. In that case the initial stage of the algorithms mentioned above will fail to match a feature belonging to repeated structures to its correct match in the second image, since it will not be able to choose the correct candidate. Thus, this feature will be discarded. In Generalized RANSAC~\citep{Jana}, all possible matches of the feature to similar (normalized cross correlation above a certain threshold) features in the second image are generated but are given low probabilities. On the list of putative matches guided RANSAC is run. Thus, in the case when there are not enough non-repeating inliers in the list with high probabilities, the algorithm might fail. In our previous work~\citep{MariaIlan12}, a special algorithm was developed to deal with buildings with repeated features. There also,  all possible matches of the feature to similar features in the second image are generated. The algorithm assumes that in both images a planar facade is visible. The algorithm tends to fail when this assumption is not satisfied. In this work we propose a method which can successfully deal with general scenes, including the case of repeated structures.

\section{Algorithm outline}\label{sect:outline}

The goal of the algorithm is to generate a set of putative feature matches between the two images, where each match is accompanied by a prior probability (or score) that the match is correct. The higher the quality of this set, the more probable that algorithms from the guided RANSAC family~\citep{PROSAC,USAC,BLOGS,BEEM} will succeed  to estimate the epipolar geometry. In this section we will present an overview of the algorithm. The details will be given in the next section.

The algorithm, given in pseudo-code in Algorithm~1, is described as follows:

\begin{algorithm}
\begin{algorithmic}[1]
\STATE Input: images $I_1$ and $I_2$
\STATE Extract SIFT features from $I_1$ and $I_2$
\STATE Find standard putative correspondences $\left\{X_L\right\}$ and associate to them distance ratios $\left\{d_r\right\}$
\STATE Find standard putative correspondences $\left\{X_B\right\}$ and associate to them similarity weights $\left\{t_k\right\}$
\STATE Cluster SIFT features from each image based on descriptor similarity yielding clusters of features
\STATE Estimate relative roll angle $\alpha_{exp}$
\FOR  {$\alpha$ $\in$ $\left[\alpha_{exp},0^{\circ}\right]$}
\STATE Match clusters from the two images, yielding cluster pairs
\STATE Generate putative correspondences $\left\{X\right\}$ from the members of the matched clusters
\STATE Generate all 2keypoints: a pair of features from a main feature point and another feature point which is close to it in the image
\STATE Match 2keypoints from the first image to the 2keypoints from the second image
\STATE Use a classifier to assign probabilities to 2keypoint matches
\STATE Select the top $K_{2kp}$ of 2keypoint matches
\STATE Estimate a candidate fundamental matrix from each two matched 2keypoints, yielding $K_{2kp}(K_{2kp}-1)/2$ matrices
\STATE For each putative match from $\left\{X\right\}$ count how many candidate fundamental matrices ($sfm$) support it 
\STATE Assign each putative match from $\left\{X\right\}\bigcup\left\{X_L\right\}\bigcup\left\{X_B\right\}$ a probability that it is correct
\STATE Use these putative matches and their associated probabilities as input to one of the algorithms from the guided RANSAC family to yield a fundamental matrix and its support
\ENDFOR
\STATE Choose from the two fundamental matrices, the one with maximal support
\RETURN The fundamental matrix and the list of its inliers
\end{algorithmic}
\label{alg:algorithm}
\caption{A General Preprocessing Method for Improved Performance of Epipolar Geometry Estimation}
\end{algorithm}

The algorithm is given as input two images $I_1$ and $I_2$. The first step of the algorithm (described in Section~\ref{sect:standard}) is to detect features (SIFT in our case) in each image. Those features are used to generate three groups of putative correspondences.

Following the standard method introduced by~\citet{SIFT}, we find putative correspondences $\left\{X_L\right\}$ and associate to them distance ratios $\left\{d_r\right\}$. The distance ratio is used to assign a prior probability for the correctness of the match (described in Section~\ref{sect:standard}).

Following the scheme introduced by~\citet{BLOGS}, we find putative correspondences $\left\{X_B\right\}$ and associate to them similarity weights $\left\{t_k\right\}$.

To resolve problems which occur in image pairs which are hard to match, such as scenes which include repeating elements, we cluster detected features (described in Section~\ref{sect:clustering}). Thus, repeated features or features with very similar descriptors are clustered together. Non-repeating features will belong to clusters of size one. Then each cluster from the first image is matched to the most similar cluster in the second image and vice versa. The result of this step is a large number of putative correspondences $\left\{X\right\}$ (described in Section~\ref{sect:keypoint_matches}). A vast majority of them however, are incorrect. 	
	
\begin{figure}[htb]
	\begin{center}	
   	\includegraphics[width=0.9\linewidth]{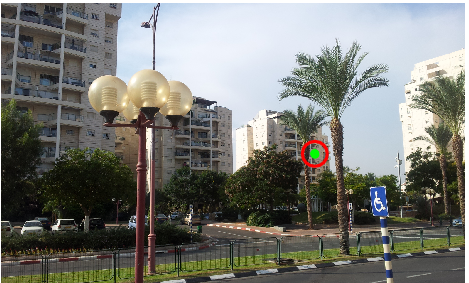}\\
   	\includegraphics[width=0.9\linewidth]{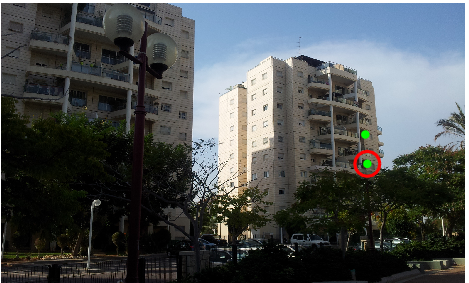}
	\end{center}
  \caption{An example of why using the standard putative matches as is, is sometimes insufficient, while clustering might help. Images GEO00038 and GEO00029 were taken from the Urban dataset. The feature point marked by a Red circle in the upper image is not matched at all by Lowe or mistakenly matched by BLOGS to a feature point marked by a Red circle in the bottom image. When using clustering, for feature point marked by the Red circle in the upper image we generate two putative matches marked by Green points in the bottom image, one of which is correct.}
\label{fig:why_clustering}
\end{figure}

An example of how clustering similar features can help in the case of a scene which includes repeating elements, is shown in Figure~\ref{fig:why_clustering}. For the feature point marked by a Red circle in the upper image~\citet{SIFT} finds no match and~\citet{BLOGS} find the feature point marked by a Red circle in the bottom image, which is incorrect. When using clustering, as we suggest, the feature point in the upper image belongs to a cluster of size one, while in the bottom image two feature points are clustered together, marked by Green points. Only in the case of clustering, a correct match is generated, namely the correct match is a member of $\left\{X\right\}$, but not of $\left\{X_L\right\}$ or $\left\{X_B\right\}$.

In addition, in order to overcome the problem of features looking similar to rotated features (such as window corners), for the clustering step only, the orientation of all the SIFT features in the image is fixed in one specific direction. Thus, for example different corners of a window will not be clustered together. In order to determine this direction, we propose to find a rough relative roll angle $\alpha$, between the two images, from the differences of SIFT orientations in $\left\{X_L\right\}$ and $\left\{X_B\right\}$ and denote it $\alpha_{exp}$. Using this approximation and the fact, that many images are taken with zero roll angle as a prior, all the following steps of our algorithm are repeated twice, once for $\alpha=\alpha_{exp}$ and again for $\alpha=0^{\circ}$.

\begin{figure*}[tb]
	\begin{center}
		\begin{tabular}{ c c c c}
   	\includegraphics[width=0.23\linewidth]{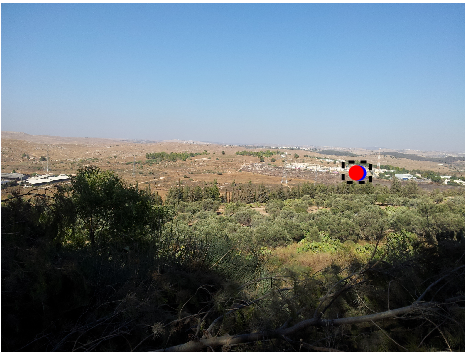}&
   	\includegraphics[width=0.23\linewidth]{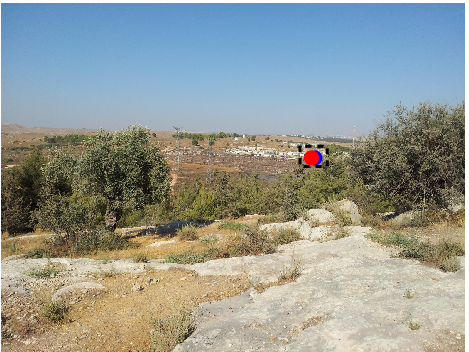}&
   	\includegraphics[width=0.23\linewidth]{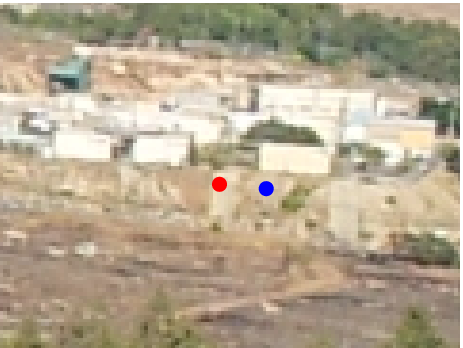}&
   	\includegraphics[width=0.23\linewidth]{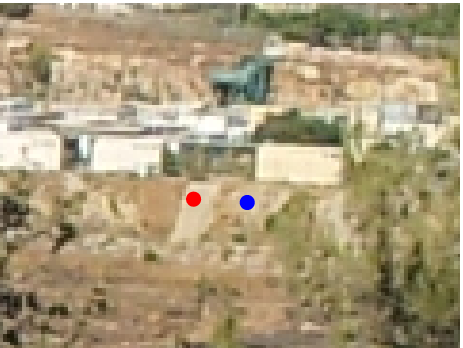}\\
   	(a)& (b)& (c)& (d)
   	\end{tabular}
	\end{center}
  \caption{An example of the advantage in using 2keypoints. Images (a) FLH00010 and (b) FLH00016 taken from the Open2 dataset. (c) Zoom-in of (a). (d) Zoom-in of (b). A correct 2keypoint match generated by our method and ranked in the fifth place. When single keypoint matches are used both Blue and Red matches are ranked much lower.}
\label{fig:figure_why_2keypoint}
\end{figure*}

In order to overcome the problem that the majority of the matches in $\left\{X\right\}$ are incorrect we estimate their probabilities to be inliers. This is done in two steps:

In the first step, which is described in Section~\ref{sect:local}, local information is used. We create a pair of features from a main feature point and another feature point which is close to it in the image. This pair of features is called a 2keypoint. These two features are matched to corresponding features in the second image which are also close to each other and belong to matching clusters. An illustration of a 2keypoint match is shown in Figure~\ref{fig:figure_why_2keypoint}.
	
The decision to work with 2keypoints is a compromise between two contradictory preferences: on the one hand any combination of features contains more information than a single keypoint, which can be used to detect inliers more accurately. In general, the larger the number of features in the combination, the higher the probability that the matched combination is correct. On the other hand, since the probability for feature detection is low, the probability for detecting a large number of features in a combination is even lower. Thus, relying on the minimal subset of features is preferable due to the difficulties in detecting large combinations of features.

From the set of 2keypoint matches we would like to choose a subset, which have a high prior probability to be correctly matched. In order to accomplish this, each 2keypoint match is characterized by a short descriptor. The descriptor consists of measures of geometric similarity between the two 2keypoints and a count of the number of possible matches between each 2keypoint in one of the images to 2keypoints in the other image. As the interdependencies between these characteristics are complex, a classifier is trained to learn the probability that the 2keypoint is correctly matched. At test time each 2keypoint match is assigned a probability and the top $K_{2kp}$ (100 in our application) 2keypoint matches are selected.

An example of how the generation and ranking of 2keypoint matches can help in dealing with low ranking matches, is shown in Figure~\ref{fig:figure_why_2keypoint}. In that example if only single keypoints are used, standard techniques~\citep{SIFT}~and~\citep{BLOGS} would order the Blue match in the 55th and 16th places respectively, whereas the Red match would be placed in a 161th place or would not be generated at all. On the other hand, when 2keypoint matches are used, the correct 2keypoint is ranked in the fifth place.

In the second step, described in Section~\ref{sect:global}, global information is used. Up until now the analysis we performed has been local in nature. We first matched single features and then pairs of close features. In order to be able to assign more accurate probabilities to the matches, the epipolar geometry constraint, which is global in nature, comes into play. In order to generate rough estimations of the fundamental matrix we borrow an idea from the BEEM algorithm~\citep{BEEM}, where it is estimated from only two matches. In our case from each two matched 2keypoints a candidate fundamental matrix is estimated, yielding $K_{2kp}(K_{2kp}-1)/2$ matrices. As a result of the ranking of the 2keypoints, quite a few of them are generated from inlier matches. The problem is that  even in this case they are quite inaccurate. Each of them is supported (i.e., the Sampson distance computed from the matrix and the match is below a certain threshold) by a subset of the inliers and quite a few outliers. Instead of returning the matrix with the largest support, we exploit these matrices in a different way. For each putative match from $\left\{X\right\}$ we count how many candidate fundamental matrices ($sfm$) support it. This number is a strong indication of the probability that this putative match is inlier.
	
Finally we combine the three groups of putative correspondences $\left\{X\right\}$, $\left\{X_L\right\}$ and $\left\{X_B\right\}$, in order to achieve a set of putative feature matches between the two images, where each match is accompanied by a prior probability that the match is correct (described in Section~\ref{sect:comb}). For that purpose we construct a keypoint match descriptor, denoted $kpmd$, and train a classifier on it. The descriptor consists of the local measures of similarity, namely $\left\{d_r\right\}$ and $\left\{t_k\right\}$ and the global measure $\left\{sfm\right\}$. At test time each putative feature match is assigned a probability.

As was already mentioned, all the previous steps of our algorithm are repeated twice, once for $\alpha=\alpha_{exp}$ and again for $\alpha=0^\circ$.
In order to proceed we run one of the algorithms from the guided RANSAC family twice, once for each set of putative matches, and choose the one with maximal support.

The output of the entire process is a fundamental matrix along with its inlier set.

\section{Algorithm details}\label{sect:detailed}
We will now delve into the details of the various components of the algorithm.

\subsection{Extraction of putative matches with their distance ratios and similarity weights}\label{sect:standard}
The algorithm is given as input two images. As a first step we apply feature detection on both images. In general, any feature detector which returns the location, scale, and orientation can be used (e.g.: MSER~\citep{MSER}, BRISK~\citep{BRISK}, ORB~\citep{ORB}, SURF~\citep{SURF}, SIFT~\citep{SIFT}). In our case we use the implementation of SIFT by~\citet{VFLEAT}.

Following the standard method introduced by~\citet{SIFT}, we find putative correspondences $\left\{X_L\right\}$ based on descriptor similarity. The best candidate match for each keypoint in the first image, is found by identifying its nearest neighbor in the second image. The nearest neighbor is defined as the keypoint with maximal normalized cross-correlation from the given descriptor vector. The probability that a match is correct can be determined by taking a distance ratio
$$d_r=\frac{\cos^{-1}(m_k)}{\cos^{-1}(m_{k_2})},$$
where $m_k$ is the similarity to the closest neighbor and $m_{k_2}$ is the second highest similarity in the second image. All matches for which the distance ratio is greater than a certain threshold (in our case 0.9) are rejected. This choice of threshold distance ratio is relatively high (there are many works where 0.85 or even 0.8 are used) and many more matches are kept. This is done since we rely on the next steps of our method to deal with them correctly (See for example Figure~\ref{fig:figure_why_2keypoint}.).

In addition we follow the scheme introduced by~\citet{BLOGS}, which is a different way to define putative correspondences and weights. We find putative correspondences $\left\{X_B\right\}$ that exhibit the highest similarity measure in both images. This means that putative correspondence $x_k = (u_k,v_k)$ is a member of $\left\{X_B\right\}$ if for a keypoint in the first image $u_k$ its nearest neighbor in the second image is $v_k$, and for the keypoint in the second image $v_k$ its nearest neighbor in the first image is $u_k$. With each such putative match pair, they associate a confidence measure which is referred to as the similarity weight. They define the similarity weight $t_k$ for the correspondence $x_k$ as
$$t_k=\left(1-\exp^{-m_k}\right)^2\left(1-\frac{m_{k_1}}{m_k}\right)\left(1-\frac{m_{k_2}}{m_k}\right),$$
where $m_{k_1}$ is the second highest similarity in the first image and $m_{k_2}$ is the second highest similarity in the second image as mentioned above. While the third term of the $t_k$, i.e. $\left(1-\frac{m_{k_2}}{m_k}\right)$ can be interpreted as another version of $d_r$, the two other terms are new. The second term, i.e. $\left(1-\frac{m_{k_1}}{m_k}\right)$ is a symmetric complimentary of the third one, and it emphasizes that there should be no difference between the treatment of first and second image. The first term, i.e. $\left(1-\exp^{-m_k}\right)^2$ is a similarity based component. While Lowe in his work did not use the absolute distance/similarity as a measure of similarity, in BLOGS the contribution of the absolute similarity exists.

\subsection{Feature clustering}\label{sect:clustering}
Using the former putative matches as is, is sometimes insufficient due to the following two problems which occur in challenging image pairs. In scenes which include repeating elements (such as for example windows of buildings), the matching process is unable to match the repeated features correctly. In image pairs with wide baselines, the descriptors of the matching features are quite dissimilar and will receive quite low matching scores. We will now deal with the first problem. The second problem will be addressed in Section~\ref{sect:local}. 

In each image, the features recovered from it in Section~\ref{sect:standard}, are clustered based on descriptor similarity. In our algorithm we use agglomerative clustering. The merging of clusters stops when the similarity measure between the closest clusters is below a certain threshold (normalized cross-correlation below 0.85). The result of this process is a set of clusters of features. Non-repeating features yield clusters of size one. Each cluster is represented by the median descriptor of its members.

\begin{figure}[htb]
	\begin{center}	
   	\includegraphics[width=0.95\linewidth]{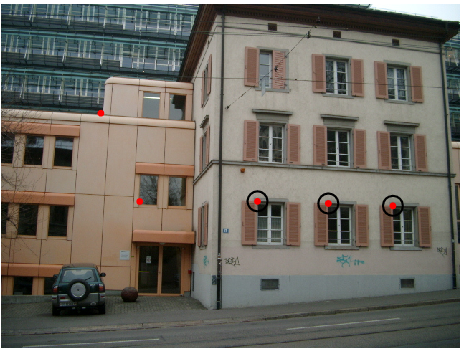}
	\end{center}
  \caption{An example of keypoint clustering with and without fixed orientation. Image object0008.view04 was taken from the ZuBuD dataset. Black circles: clustering with fixed SIFT orientation. Red points: clustering without fixed orientation. In the latter case three different types of corners are clustered together.}
\label{fig:Clustering_with_angles_example}
\end{figure}

\begin{figure*}[tb]
	\begin{center}
		\begin{tabular}{ c c c}
   	\includegraphics[width=0.3\linewidth]{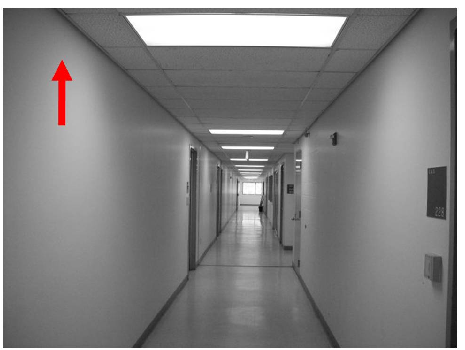}&
   	\includegraphics[width=0.3\linewidth]{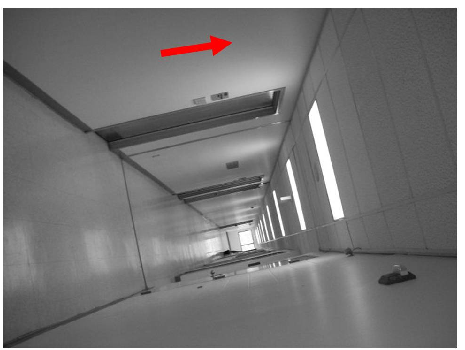}&
   	\includegraphics[width=0.3\linewidth]{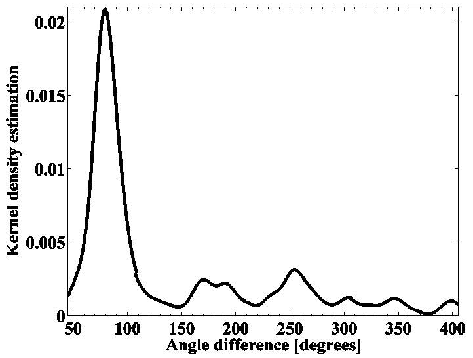}\\
   	(a) & (b) & (c)\\
   	\includegraphics[width=0.3\linewidth]{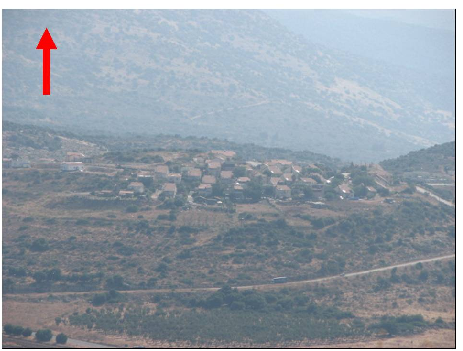}&
   	\includegraphics[width=0.3\linewidth]{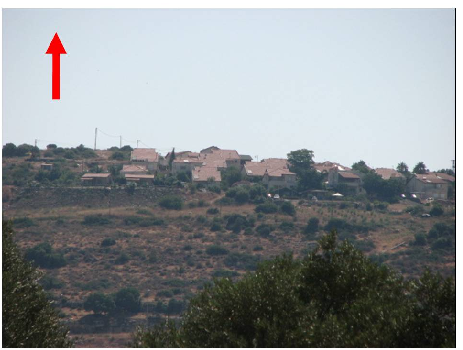}&
   	\includegraphics[width=0.3\linewidth]{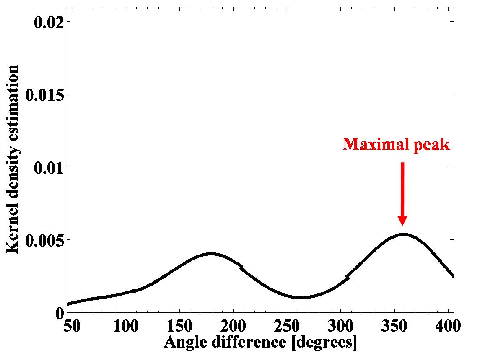}\\
   	(d) & (e) & (f)\\
   	\end{tabular}
	\end{center}
  \caption{An example of a relative roll angle $\alpha$ estimation. Images (a) corridor1 and (b) corridor2 were taken from the BLOGS dataset. The images in (a) and (b) are taken with relative roll of $\alpha=78^{\circ}$. (c) The kernel density estimation of the angle difference for (a) and (b) with the maximal peak at $\alpha_{exp}=78^{\circ}$. Images (d) IMG0047 and (e) IMG0106 were taken from the Open1 dataset. The images in (d) and (e) are taken with zero relative roll. (f) The kernel density estimation of the angle difference for (d) and (e) with the maximal peak at $\alpha_{exp} = -3^{\circ}$.}
\label{fig:roll_extraction}
\end{figure*}

In order to overcome the problem, which is common in buildings, of features looking similar to rotated features (such as window corners), for the clustering step only, the orientation of all the SIFT features in the image is fixed in one specific direction. Thus, for example different corners of a window will not be clustered together. An example of keypoints clustering with and without fixing the orientation of all the SIFT features is presented in Figure~\ref{fig:Clustering_with_angles_example}. The Red points show a cluster without fixed orientation. In that case three different orientations of the window corner are clustered together. The Black circles are features clustered together, when fixing the SIFT orientation. All of them are upper left corners of a window. In general, clustering without fixed orientation of all the SIFT features in the image, leads not only to larger clusters which can be handled by our method, but to systematic errors when matching features from those clusters. This will be further explained later on.

The approach, of defining one specific orientation for all the SIFT features in the image, has been extensively used in the literature in the frame of upright SIFT, and in this work we generalize this idea to any orientation. It is true, that there are many applications such as vision based robot navigation and structure from motion, where all the images are taken with a zero roll angle, which justifies the upright SIFT assumption in all the images. However, since we do not limit our approach to any specific application, we propose to find a rough approximation of the relative roll angle $\alpha$, between the two images, from the existing data. For that purpose we calculate the difference of SIFT orientations in each putative match found in Section~\ref{sect:standard}, and build a kernel density estimation of those angle differences. Although the transformation between the two images is perspective and not affine, the maximal peak $\alpha_{exp}$ of this function can be used as a rough approximation for $\alpha$.

Two examples of $\alpha$ extraction are shown in Figure~\ref{fig:roll_extraction}. The red arrows were added to indicate the upright direction. In the first row the image pair, taken with a relative roll of $\alpha = 78^{\circ}$, along with its kernel density estimation of the angle difference is presented. The maximal peak of this kernel density is precisely $\alpha_{exp} = 78^{\circ}$. In the second row an image pair, taken with zero relative roll is shown. The maximal peak of its kernel density estimation is located at $\alpha_{exp} = -3^{\circ}$ which is quite a good approximation for $\alpha$.

Using this approximation and the fact that many images are taken with zero roll angle as a prior, we proceed as follows. When running the algorithm, the first image is processed one time using an upright SIFT, while the second image is processed in two orientations $\left[\alpha_{exp},0^{\circ}\right]$. Therefore all the following steps of our algorithm, described in Sections~\ref{sect:keypoint_matches}-\ref{sect:comb} are repeated twice, once for each orientation.

%In addition, we have found out empirically that there are common cases, especially in urban scenes containing straight lines, when there are two almost equal peaks in the histogram of angle differences, spaced by $180^{o}$.

\subsection{Generation of all keypoint matches}\label{sect:keypoint_matches}
After both of the images have been processed as described above, the next step is to generate pairs of possible matches between features from the two images. The main problem we have to overcome is when a real cluster is segmented into several clusters. We try to deal with this problem as follows. Each cluster from the first image is matched to the closest cluster from the second image using the normalized cross-correlation between the cluster representatives. This process is repeated when the roles of the images are switched. Thus, if a real cluster was over-segmented in one image but not in the other we can still match the clusters from the two images correctly. If there is over-segmentation in both images, not all possible matches will be found.

Due to this problem we can not use the distance ratio method suggested by~\citet{SIFT}. Because if the distances to the closest cluster and the second closest cluster are similar, we can not distinguish between  over-segmentation and  when  both clusters in the second image equally far from the cluster in the first image and should not be matched at all. Therefore the closest cluster is always chosen.

\begin{figure}[htb]
	\begin{center}	
		\begin{tabular}{ c c }
   	\includegraphics[width=0.45\linewidth]{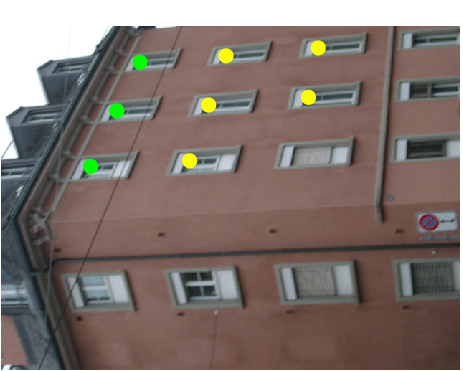}&
   	\includegraphics[width=0.45\linewidth]{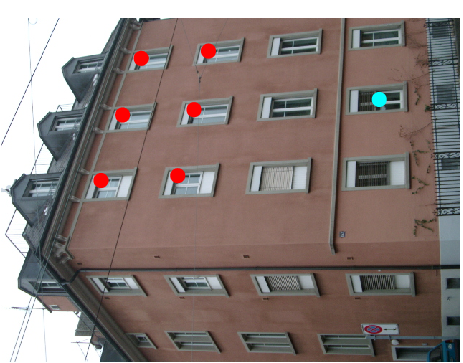}\\
   	(a) & (b)\\
   	\end{tabular}
	\end{center}
  \caption{An example of keypoint clustering and matching. Images (a) object0076.view02 and (b) object0076.view04 were taken from the ZuBuD dataset. The large Red cluster in (b) is segmented into two smaller (Green and Yellow) clusters in (a). In this example due to cluster matching in both directions all the correct matches were found.}
\label{fig:Clustering_example}
\end{figure}

An example of the result of the clustering process is shown in Figure~\ref{fig:Clustering_example}. The large Red cluster in the second image is segmented into two smaller (Green and Yellow) clusters in the first image. The Red cluster was matched to the Yellow cluster when clusters from the second image are matched to clusters in the first image. Because the matching is also done from clusters in the first image to the closest cluster in the second image, the Green cluster is also matched to the Red cluster. In this example all of the correct matches were found together with many incorrect matches.

Each pair of clusters which is matched yields a set of putative feature matches from the members of the two clusters. The result of this step is a large number of possible matches most of which are obviously incorrect.

We will now refer to the systematic errors mentioned above when explaining the clustering without a fixed orientation of all the SIFT features in the image. As was already shown, when the orientation of all the SIFT features is not kept fixed, features looking similar to rotated features (such as window corners) will be clustered together. In that case, when generating putative feature matches, there will be matches of the same feature (left upper corner in both images for example) and there also will be matches of the rotated features (left upper corner in the first image matched to right lower corner in the second image for example). In the next steps of the algorithm, these matches of the rotated features will all vote together supporting each other and will lead to systematic errors. We therefore chose to keep the orientation of all the SIFT features fixed during the clustering in Section~\ref{sect:clustering}, to prevent such failures.

\subsection{Generation and ranking of 2keypoint matches}\label{sect:local}
Most of the feature pairs generated in the previous stage are incorrect and therefore in order to be able to use them for epipolar geometry estimation  prior probabilities have to be assigned to them. This will be done in two steps: a step which uses only local information which will be described here and a step which uses global information described in the next section.

\begin{figure*}[bt]
	\begin{center}
		\begin{tabular}{ c c c c}
   	\includegraphics[width=0.22\linewidth]{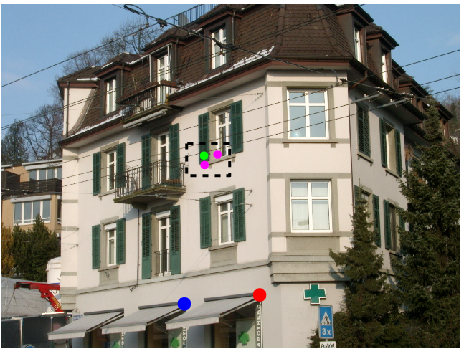}&
   	\includegraphics[width=0.22\linewidth]{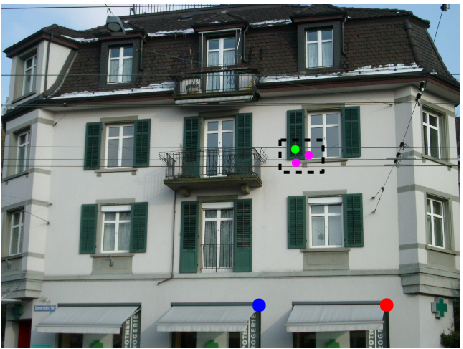}&
   	\includegraphics[width=0.22\linewidth]{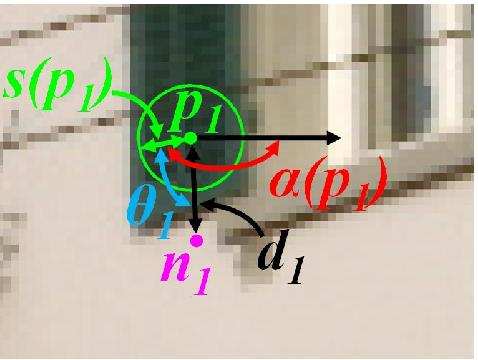}&
   	\includegraphics[width=0.22\linewidth]{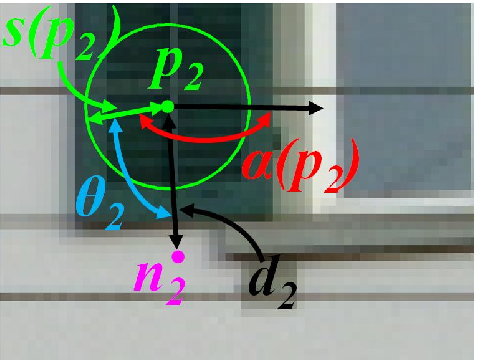}\\
   	(a) & (b) & (c) & (d) \\
   	\end{tabular}
	\end{center}
  \caption{An example of 2keypoints and the parameters used for their matching. Images (a) object0041.view01 and (b) object0041.view04 were taken from the ZuBuD dataset. (c) Zoom-in of (a). (d) Zoom-in of (b). Green and Magenta points show the 2keypoints generated by the first method. Blue and Red points show a 2keypoint generated by the third method.}
\label{fig:2keypoints_example}
\end{figure*}

Recall that each SIFT feature $p$ has besides a descriptor also a scale $s(p)$ and an orientation angle $\alpha(p)$. These values will be used in our analysis to make it scale and orientation invariant. For each feature point $p$ we add a neighboring feature $n$. The distance between the features in terms of the scale of $p$ is denoted $d = |p-n|/s(p)$. The angle between  the vector connecting  $p$ to $n$ with respect to $\alpha(p)$ is denoted $\theta$. This pair of features will be termed a 2keypoint. A 2keypoint $\{p_1,n_1\}$ in the first image is matched to a 2keypoint $\{p_2,n_2\}$ in the second image. Naturally, $p_1$ and $p_2$, and  $n_1$ and $n_2$, have to be putative matches.
This set of four features is illustrated in Figure~\ref{fig:2keypoints_example}.

We suggest three methods to choose the neighboring pairs $\{n_1,n_2\}$ close to $\{p_1,p_2\}$. The first method simply takes $n_i,\,\,i=1,2$  from the $K_1$ closest features around $p_i$. The second method chooses $n_i$ from all the features within a certain distance $K_2 s(p_i)$ in pixels from $p_i$. This parameter is given  in units of scale in order to be scale invariant. Finally, the third method chooses $n_i$ from the  $K_3$ closest features which belong to the same cluster as $p_i$. Experimentally we found that optimal values are achieved for $K_1=5, K_2=5,$ and $K_3=1$.

In order to estimate the probability that a 2keypoint match consists only of inliers, we have to take into account quite a few factors. Due to the interdependencies  between these factors and their effects on the estimated probability, we construct a 2keypoint match descriptor, denoted $2kpmd$, and train a classifier on it.

The descriptor consists of the following fields:

$$2kpmd=[N_1;N_2;dist_r;angle_d;cluster_t;min_d].$$

The definitions of the fields are as follows:
$N_1$  and $N_2$  are the number of 2keypoint matches that the 2keypoints $\{p_1,n_1\}$ and $\{p_2,n_2\}$ belong to respectively. The smaller the values,  the higher the probability that the 2keypoint match consists of inliers.
$dist_r = \min(d_1/d_2,d_2/d_1)$ is the ratio of the distance between $p_1$ and $n_1$ in the first image in terms of $s(p_1)$ to the distance between  $p_2$ and $n_2$ in the second image in terms of $s(p_2)$. This measure is scale invariant and its value should be close to one.
The value $angle_d = ang_{diff}(\theta_1,\theta_2)$ measures the difference between the angles associated with the two 2keypoints and should be close to zero. The field $cluster_t$ is equal one if $p_i$ and $n_i$ belong to the same cluster  and zero otherwise. Finally, the distance between the point and its neighbor also affects the probability that the 2keypoint match is correct. The further the points are, the lower the probability is. We therefore define $min_d=\min(d_1,d_2)$.

\begin{figure}[htb]
	\begin{center}
   	\includegraphics[width=0.95\linewidth]{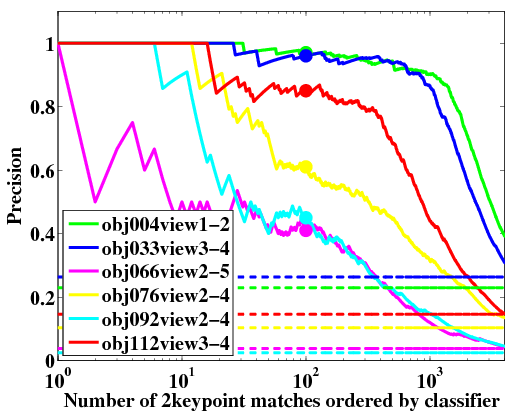}
	\end{center}
  \caption{2keypoint match classifier performances on the training set. Solid curves represent results of the classifier. Dashed curves show results without the classifier (the inliers percentages from all the 2keypoint matches). When using the classifier, there are many more inliers in the top $K_{2kp}=100$ 2keypoint matches.}
\label{fig:2keypoints_classifier}
\end{figure}

In order to train the classifier we chose six image pairs from the ZuBuD dataset. From them four were cases that state-of-the-art algorithms were able to match, while the other two were more challenging. For these image pairs we manually found the ground truth matches and trained on the data a C4.5 decision tree classifier~\citep{C4.5}. This classifier returns for each descriptor the probability that the 2keypoint match consists of inliers.
The training set consists of 31352 2keypoint matches of which 4102 are inliers and the rest are outliers. The quality of the classifier was estimated using a 10-fold cross-validation procedure. When choosing a classifier, we tried several options such as random forest~\citep{RandomForest}, SVM~\citep{SVM} and others. The C4.5 decision tree classifier was selected, as the one which not only classifies correctly 91.6\% of the 2keypoint matches, but also gives a maximal precision (the proportion of positive results that are true positive) of 73.8\%, which is the most important parameter as will be now explained.

The 2keypoint matches are then sorted by probability and the highest $K_{2kp}$ ($K_{2kp}=100$ in our implementation) are chosen.
Thus, what is most important is that from the top $K_{2kp}$ a fair amount of them should be inliers (precision). This is evident from the results on the training set shown in Figure~\ref{fig:2keypoints_classifier}. The cumulative precision of the classifier is shown as a function (on a log scale) of the number of 2keypoint matches ordered by the probability returned by the classifier. Dashed curves represent the inlier percentages from all the 2keypoint matches, which would be correct if no classifier existed. Consider for example the hardest case of the image pair (obj066view2,obj066view5). From the top ranked 100 2keypoint matches, 41\% were inliers, while their percentage from all the 2keypoint matches was only 3.84\%.

\subsection{Global ranking of matches}\label{sect:global}
Even though we could use the 2keypoint matches found in the previous step as the input for epipolar geometry estimation, better results can be obtained by exploiting global information. In BEEM~\citep{BEEM} a method was proposed to generate a rough estimate of the fundamental matrix using only two pairs of matches instead of 7 or 8. This is done by using the similarity transformation between the regions around the corresponding features, to generate three additional matches for each ``real'' match. The resulting estimated fundamental matrix is quite inaccurate but can be used as a basis for local optimization~\citep{LO_RANSAC}, yielding good results. In our case we use two 2keypoint matches (four matched points) as the input for estimating the fundamental matrix. One 2keypoint match could not be used since all the points from each  image are too close to each other to generate a meaningful result.

For  each of the $K_{2kp}(K_{2kp}-1)/2$ pairs of 2keypoint matches a fundamental matrix $F$ is generated. All the putative matches generated in Section~\ref{sect:keypoint_matches} are checked to see whether they support $F$ or not. Instead of taking the fundamental matrix with the largest support as the result of our algorithm, we suggest here a method to exploit all the generated fundamental matrices. Since we assume that many of them were generated from inlier 2keypoint matches they are therefore rough estimates of the required solution. Thus, we measure the support of the putative matches. The larger the number of fundamental matrices which support the match ($sfm$), the higher the probability that the match is correct.

\begin{figure}[htb]
	\begin{center}
		\includegraphics[width=0.95\linewidth]{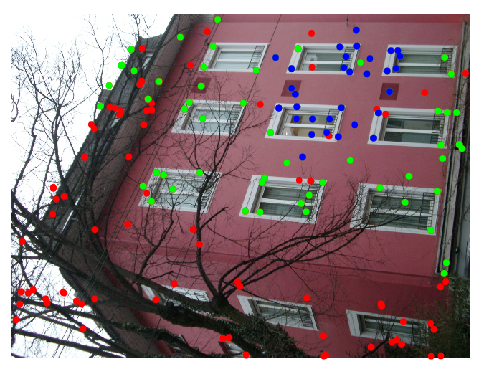}
	\end{center}
  \caption{An example of most supported inliers. Image object0092.view02 was taken from the ZuBuD dataset. Matches supported by more than 600 $sfm$s are plotted in Blue, matches with more than 200 $sfm$s in Green, and the rest in Red.
\label{fig:keypoints_matches_supported_by_Fs}}
\end{figure}

An example of the spatial distribution of the inlier matches is shown in Figure~\ref{fig:keypoints_matches_supported_by_Fs}. Since the fundamental matrices generated from inliers are quite inaccurate, only a small number of matches which lie close to each other, are supported by a large number of fundamental matrices (marked in Blue). The other matches with lower support are distributed around this group in an irregular manner.

\begin{figure}[htb]
	\begin{center}
   	\includegraphics[width=0.95\linewidth]{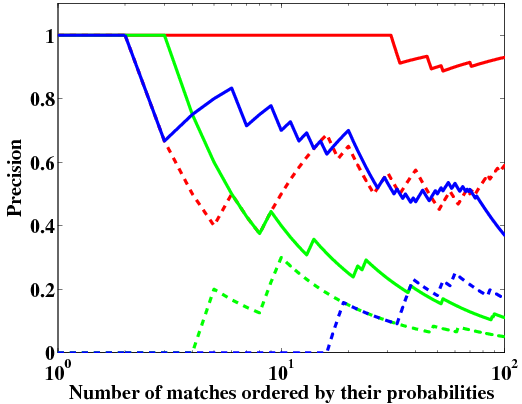}
	\end{center}
  \caption{2keypoint match classifier performance vs. global ranking of matches performance. Red: images object0076.view02 and object0076.view04 were taken from the ZuBuD dataset, shown in Figure~\ref{fig:Clustering_example}. Green: images FLH00010 and FLH00016 were taken from the Open2 dataset, shown in Figure~\ref{fig:figure_why_2keypoint}. Blue: images GEO00029 and GEO00038 were taken from the Urban dataset, shown in Figure~\ref{fig:why_clustering}. Solid curves show the results of exploiting global information, while the dashed curves show the results found in Section~\ref{sect:local}. Better results are obtained by exploiting global information.}
\label{fig:how_global_ranking_improves}
\end{figure}

In Figure~\ref{fig:how_global_ranking_improves} we compare results obtained by exploiting global information, to the ones obtained using only the 2keypoint matches found in Section~\ref{sect:local}. For that, we present the cumulative precision (inliers' fraction) as a function of the number of putative matches, ordered based on their associated probabilities. The solid curves show the results obtained by exploiting global information, namely cumulative precisions as a function of number of keypoint matches ordered by their $sfm$s. The dashed curves show the results found in Section~\ref{sect:local}, namely the cumulative precision as a function of the number of keypoint matches ordered by the probability returned by the 2keypoint match classifier. Each color represents a different image pair. As can be seen in the graph, better results are obtained by exploiting global information, in addition to using the 2keypoint match ranking computed in Section~\ref{sect:local}. Consider for example the hardest case of the image pair (FLH00010,FLH00016). From the top ranked 100 putative matches ordered by their $sfm$s, 11\% were inliers, while if they were ordered by the probability returned by the 2keypoint match classifier only 5\% would be inliers.

The result of this step is the set of putative matches $\left\{X\right\}$ described in Section~\ref{sect:keypoint_matches} and their $sfm$s $\left\{sfm\right\}$.

\subsection{Combining all the data}\label{sect:comb}
As was stated earlier, the goal of the algorithm is to generate a set of putative feature matches between the two images, where each match is accompanied by a prior probability (or score) that the match is correct. For that purpose, in Sections~\ref{sect:clustering}-\ref{sect:global} we presented a three step algorithm, running from local to global and generating putative matches and their $sfm$s. In addition, in Section~\ref{sect:standard}, we calculated putative match pairs $\left\{X_L\right\}$ and $\left\{X_B\right\}$, which have a large intersection with $\left\{X\right\}$, along with their distance ratios $\left\{d_r\right\}$ and/or a similarity weights $\left\{t_k\right\}$, based on the local features only. In order to incorporate those local scores in our method, we constructed a keypoint match descriptor, denoted $kpmd$, and trained a classifier on it.

The descriptor consists of the following fields:
$$kpmd=[sfm;d_r;t_k].$$
The definitions of the fields are as follows: $sfm$ is the number of fundamental matrices which support the match, calculated in Section~\ref{sect:global}. $d_r$ and $t_k$ are the distance ratio and the similarity weight described in Section~\ref{sect:standard} respectively. For those putative matches that miss $d_r$ or $t_k$, we attribute ones for $d_r$, and zeros for $t_k$. For those putative matches in $\left\{X_L\right\}$ and $\left\{X_B\right\}$ that miss the $sfm$ we attribute zeros. In general, the smaller the value of $d_r$ and the higher the values of $sfm$ and $t_k$, the higher is the probability that the putative feature match is an inlier.

The general idea behind this step is to improve the performance on challenging image pairs, while not harming the performance on easy ones. For that purpose the classifier should  operate correctly under different scenarios. On the one hand, when an image pair is challenging, putative match pairs $\left\{X_L\right\}$ and $\left\{X_B\right\}$ are insufficient and it should rely on $\left\{X\right\}$ and their $sfm$s. On the other hand, for easy image pairs $\left\{X\right\}$ might be misleading, while relying on $\left\{X_L\right\}$ and $\left\{X_B\right\}$ works. Since there is no way to know a-priori with which scenario we are dealing with, the classifier should highly rank both: putative matches with high $sfm$s and missing (or low) $t_k$ and $d_r$, and match pairs with missing $sfm$s but with high $t_k$ and/or low $d_r$ values.

Using the training set described above, we trained a C4.5 decision tree classifier~\citep{C4.5}. This classifier returns for each descriptor the probability that the putative feature match is an inlier. The training set consists of 14255 feature matches from which 1399 are inliers and the rest are outliers. Here again a 10-fold cross-validation procedure was run. The resulting classifier correctly classifies 94.9\% of the feature matches.

The result of this step is a set of putative matches and their associated probabilities.

\subsection{Epipolar geometry estimation}\label{sect:F}
As was already mentioned in Section~\ref{sect:clustering}, the steps of our algorithm described in Sections~\ref{sect:keypoint_matches}-\ref{sect:comb} are repeated twice, once for each orientation. Therefore at this stage there are actually two sets of putative matches and their associated probabilities. To finalize the process we run an algorithm from the guided RANSAC family twice, once for each set of putative matches, yielding two fundamental matrices. The one with maximal support (the larger number of inliers) is chosen.

\section{Experiments}\label{sect:experiments}
Our method is a preprocessing step for state-of-the-art algorithms for epipolar geometry estimation. Therefore, in order to evaluate it, we compared the performance of three known algorithms BEEM, BLOGS and USAC with and without our method. In all the three cases we used the original implementations including all algorithm parameters, as proposed by their authors, available on the Internet. We ran experiments with the same parameters on all the
results included in this work. These parameters were automatically selected to produce optimal results.

\subsection{Test Data}\label{sect:test_data}
To demonstrate the generality of our method, we used almost 900 image pairs from six separate publicly available sources for test data. Each image pair except those from the ``USAC dataset'' came with a small set of ground truth correspondences, which are different from the SIFT features used to estimate the epipolar geometry. These correspondences were used by the authors in their performance evaluation.  The mean of roots of their Sampson distances served as our quantitative performance measure. The lower the value, the closer the proposed solution is to the ground truth.

\paragraph{ZuBuD dataset~\citep{ZuBuD}:} The dataset contains 1005 color images of 201 buildings (5 images per building) from Zurich, taken from different viewpoints and under different illumination conditions, yielding 2010 image pairs. In~\citep{MariaIlan12} two subsets of it were used: the  ``ZuBuD1 set'' of 139 challenging image pairs (two of which we used for training as mentioned in Section~\ref{sect:detailed} and the rest for test) and the  ``ZuBuD2 set'' of relatively easy image pairs. This way we can check the performance of the algorithm on both hard and easy cases.

\paragraph{BLOGS dataset~\citep{BLOGS_DB}:} The BLOGS dataset consists of 20 image pairs, some of which have very wide baselines, scale changes, rotations and occlusions.

\paragraph{USAC dataset~\citep{USAC_DB}:} In the USAC dataset there are 11 image pairs. Since image pairs from this dataset come without control points, we manually marked 16 correspondences for each image pair, serving as the ground truth.

Since the ``BLOGS dataset'' and the ``USAC dataset'' are quite small, in our experiments we merged them into a single dataset.
\paragraph{Open1, Open2 and Urban datasets~\citep{SOREPP_DB}:} These three datasets that were collected at different locations include 246, 224 and 108 image pairs respectively. They were used for testing the SOREPP algorithm~\citep{SOREPP}. The datasets present challenging scenarios with wide baseline images, small overlapping regions, scale changes, and nondescript objects that make feature matching difficult. Under these conditions the inlier fractions are often less than 10\%.

\subsection{Qualitative results}
\begin{figure}[htb]
	\begin{center}
   	\includegraphics[width=0.95\linewidth]{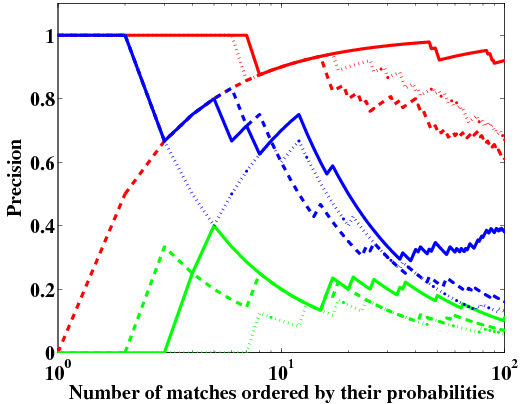}
	\end{center}
  \caption{A comparison between different methods of ranking. Red: images object0076.view02 and object0076.view04 were taken from the ZuBuD dataset, shown in Figure~\ref{fig:Clustering_example}. Green: images FLH00010 and FLH00016 were taken from the Open2 dataset, shown in Figure~\ref{fig:figure_why_2keypoint}. Blue: images GEO00029 and GEO00038 were taken from the Urban dataset, shown in Figure~\ref{fig:why_clustering}. Solid curves are results of our method. Doted curves are based on distance ratio proposed by Lowe. Dashed curves are based on similarity weights introduced in BLOGS. Our ranking method outperforms the standard ones.}
\label{fig:comparison}
\end{figure}

We will start this discussion with a presentation of qualitative results on the three image pairs already mentioned in Figures~\ref{fig:why_clustering},\ref{fig:figure_why_2keypoint},\ref{fig:Clustering_example} and \ref{fig:how_global_ranking_improves}. Figure~\ref{fig:comparison} shows the cumulative precision (inliers fraction) as a function of the number of putative matches, ordered based on their associated probabilities (on a log scale). The results of our method are drawn using solid curves. The putative matches ranking, based on the distance ratio proposed by Lowe and used as an initial step of many state-of-the-art algorithms such as USAC and BEEM, are drawn using dotted curves. The putative matches ranking, based on similarity weights introduced in BLOGS, is drawn using dashed curves. Different colors represent the different image pairs. One can easily see that our ranking method outperforms the standard ones.

\begin{table}
{\scriptsize
	\begin{center}
		\begin{tabular}{|l|c|c|c|c|}			
			\hline
			\hline			
			                   		          & Lowe/     &       & Our\\
			                   		          & USAC/BEEM & BLOGS & method\\
			\hline
			\hline
			\multicolumn{4}{|c|}{object0076.view02 and object0076.view04} \\			
			\hline
			\hline
			Number of matches			  & 415   & 355   & 5347\\
			\hline
%			Weighted precision		  &46.4\% &61.6\% & 63.25\%\\
%			\hline
			\# inliers from top 10  & 9     & 9     & 9\\
			\hline
			\# inliers from top 100 & 67    & 61    & 92\\
			\hline
			Success                 & \checkmark \checkmark &\checkmark &\checkmark \checkmark \checkmark \\
			\hline
			\hline			
			\multicolumn{4}{|c|}{FLH00010 and FLH00016} \\			
			\hline
			\hline		
			Number of matches			  & 493   &  1161 & 7735\\
%			\hline
%			Weighted precision		  & 2.8\% & 5.4\% & 1.67\%\\
			\hline
			\# inliers from top 10  & 1     & 2     & 2\\
			\hline
			\# inliers from top 100 & 7     & 7     & 11\\
			\hline
			Success                 & - -   & -     & - - -\\
			\hline			
			\hline
			\multicolumn{4}{|c|}{GEO00029 and GEO00038} \\						
			\hline
			\hline
			Number of matches			  & 674         & 1038 & 9469\\
%			\hline
%			Weighted precision		  & 4.4\%       &14.8\%&4.73\%\\
			\hline
			\# inliers from top 10  & 6           & 6    & 7\\
			\hline
			\# inliers from top 100 & 13          & 16   & 38\\
			\hline
			Success                 & - \checkmark& -    & \checkmark \checkmark \checkmark\\
			\hline
			\hline			
		\end{tabular}
	\end{center}}
	\caption{A numeric comparison of our method, on several examples, to two common techniques for ranking putative matches.}
	\label{table:comparison}
\end{table}

\begin{table*}[tb]
	\begin{center}
		\begin{tabular}{|l|c|c|c|c|c|c|}			
			\hline
			\hline			
			                   		          & ZuBuD1& ZuBuD2& BLOGS + USAC DB & Urban & Open1 & Open2\\
			\hline
			\hline
			Number of image pairs			      & 137   & 137   &  31             & 108   & 246   & 224 \\
			\hline
			\hline
			\textbf{BLOGS}                           & 69.4  & 135   & 24.4            & 35.8  & 80.2  & 52  \\
			\hline
			Our method followed by BLOGS    & 104.6 & 133.8 & 26.8            & 54.2  & 119.2 & 90  \\
			\hline
			Our contribution for BLOGS      & 50.7\%& -0.9\%& 9.8\%           & 51.4\%& 48.6\%& 73.1\%\\
			\hline
			\hline
			\textbf{BEEM}                            & 96    & 134.2 & 26.2            & 46.6  & 104.8 & 76.6 \\
			\hline
			Our method followed by BEEM     & 107   & 134.2 & 27.2            & 64    & 132.4 & 124.2\\
			\hline
			Our contribution for BEEM       & 11.5\%& 0     & 3.8\%           & 37.3\%& 26.3\%& 62.1\%\\
			\hline
			\hline
			\textbf{USAC}                            & 66    & 133.8 & 24.4            & 23.6  & 77.6  & 27  \\
			\hline
			Our method followed by USAC     & 95.6  & 132.2 & 26.2            & 51    & 124.8 & 91.6\\
			\hline
			Our contribution for USAC       & 44.8\%& -1.2\%& 7.4\%           & 116\% & 60.8\%& 239\%\\
			\hline
			\hline
		\end{tabular}
	\end{center}
	\caption{Numeric comparison of general performance with and without our preprocessing step on the standard datasets. Number of successful image pairs, with performance measure smaller than 10 pixels is reported.}
	\label{table:performance}
\end{table*}

In Table~\ref{table:comparison} we present numeric comparisons of our method on these three examples to these two common techniques for ranking putative matches. The number of matches we generate is ten times larger than the other methods. Even so, as our ranking is much better correlated with the probability to be an inlier, as was already shown in Figure~\ref{fig:comparison}, the performance is not hurt. Numerically speaking, we recover more inliers in the top 10 and top 100 ranked matches. This should be translated into improved performance of the subsequent registration process. To check that we ran BEEM, BLOGS and USAC with and without our preprocessing method and report on their success which will be defined in the next section. The first example (object0076.view02,object0076.view04) is an easy case and all the algorithms with or without the preprocessing step succeed on it. The second image pair (FLH00010,FLH00016) is so challenging, that although the number of inliers was increased by our method, it remains unsolved in all the cases. The last image pair (GEO00029,GEO00038) is a typical example of our contribution. Without our method only BEEM found a correct fundamental matrix, whereas when our preprocessing step is used, all three algorithms succeed.

\subsection{General performance}
To evaluate the general performance of our method, we present a comparison between BEEM, BLOGS and USAC with and without our preprocessing step on the previously mentioned datasets. We use ground truth correspondences and the quantitative performance measure, mentioned in Section~\ref{sect:test_data}. For every algorithm on each image pair we check this performance measure and consider it as a success when it is smaller than a threshold.

\begin{figure*}[tb]
    \centering
    \begin{tabular}{c c}
        \includegraphics[width=0.45\textwidth]{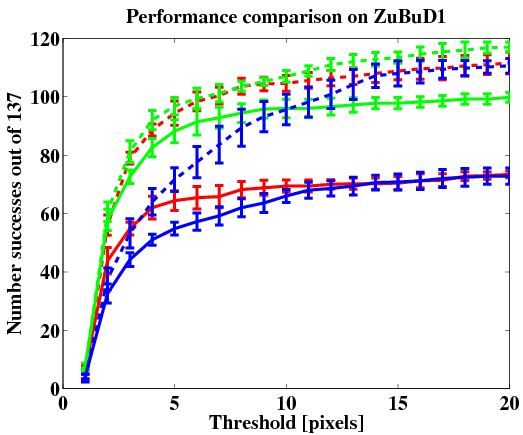}&
        \includegraphics[width=0.45\textwidth]{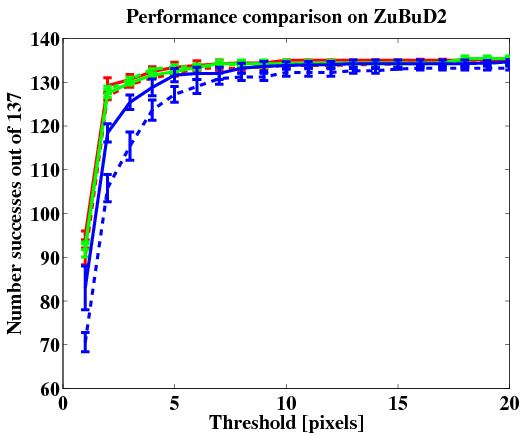}\\
        \includegraphics[width=0.45\textwidth]{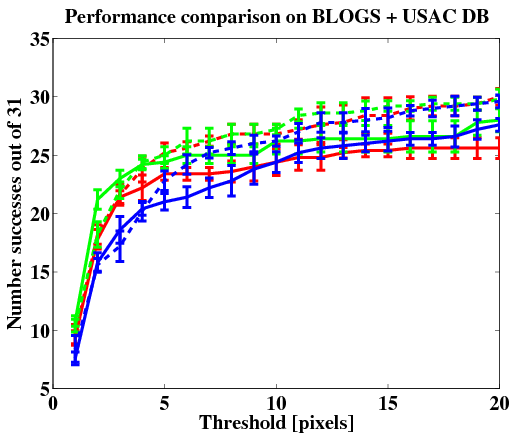}&
        \includegraphics[width=0.45\textwidth]{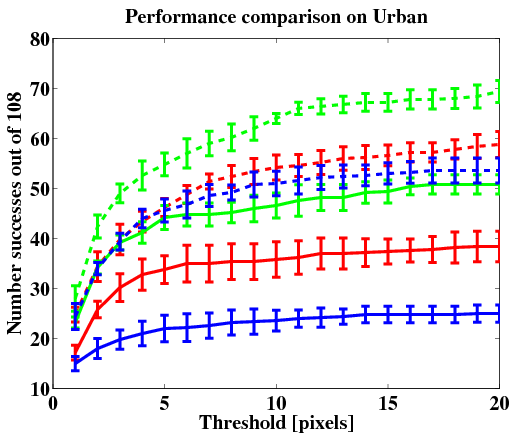}\\
        \includegraphics[width=0.45\textwidth]{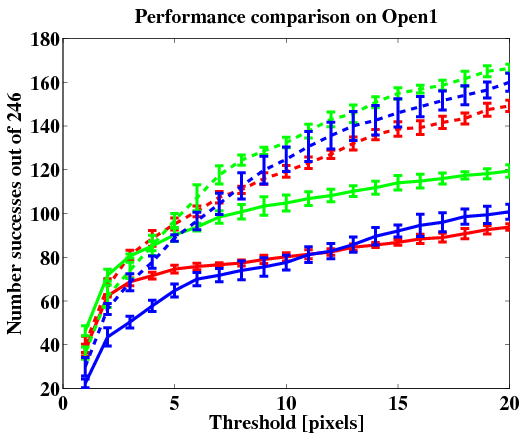}&
        \includegraphics[width=0.45\textwidth]{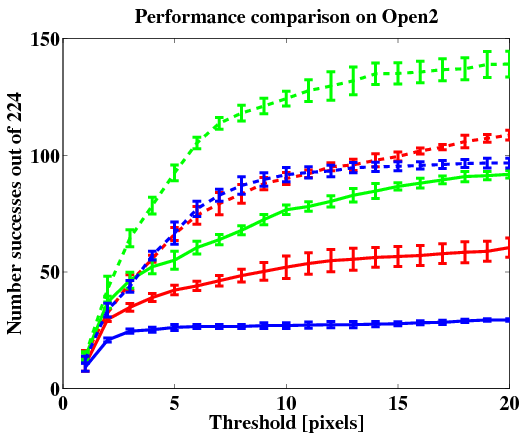}
    \end{tabular}
    \caption{Performance comparison between several algorithms with and without our preprocessing step on the standard datasets. Solid Red curves: BLOGS. Dashed Red curves: our method followed by BLOGS. Solid Green curves: BEEM. Dashed Green curves: our method followed by BEEM. Solid Blue curves: USAC. Dashed Blue curves: our method followed by USAC.}
    \label{fig:Results}
\end{figure*}

In Figure~\ref{fig:Results} we present the number of correct epipolar geometry estimations on each set of image pairs as a function of the threshold. Although our method is deterministic, algorithms for epipolar geometry estimation are not. Therefore for the sake of proper comparison, we show an average over 5 executions of the algorithms. The error bars represent one standard deviation. On the ``ZuBuD2 set'' and the ``BLOGS+USAC dataset''  all of the checked algorithms perform extremely well. The performance after our preprocessing is similar. The results for the other datasets are dramatically lower for all the checked algorithms. This indicates that many of the image pairs are challenging. The significant improvement due to our preprocessing method can be seen in all the checked algorithms. For example, for the ``Urban dataset'' and the ``Open2 dataset'' our preprocessing step improved the performance by a factor of two or three for USAC.

In Table~\ref{table:performance} we present numeric comparisons of the general performance with and without our preprocessing step. For each algorithm we report its results with and without our step, followed by our contribution for this algorithm. Our contribution is the percentage change, computed as follows:
$$ \frac{result\ with\ our\ step - result\ without\ our\ step}{result\ without\ our\ step},$$
where results with and without our step are defined as the number of successful image pairs, with performance measure smaller than 10 pixels. From all the checked cases there is a negligible degradation due to our method, of one or two out of 137 image pairs, in ZuBuD2 dataset for BLOGS and USAC. In all other verified cases, our preprocessing step improves performance of all the checked algorithms. In general we can summarize that our preprocessing algorithm yields better results in hard cases and does not degrade on the easy ones.

\begin{figure}[htb]
    \centering
    	\includegraphics[width=0.45\textwidth]{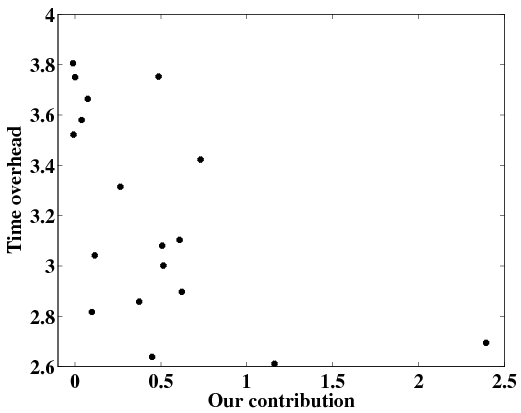}
    \caption{Time overhead of our method. There is a negative correlation between time overhead of our method and its contribution.}
    \label{fig:Time}
\end{figure}

Another issue worth mentioning is run time. Our step is a preprocessing step for any epipolar geometry algorithm. As such, the run time can not be shorter when our method is used. Moreover, as described in Section~\ref{sect:F}, it requires to run the algorithm from the guided RANSAC family twice, once for each set of putative matches. Therefore, the run time with our method is expected to be at least doubled. In Figure~\ref{fig:Time} we show its time overhead. We defined it as:
$$ \frac{run\ time\ with\ our\ step}{run\ time\ without\ our\ step},$$
and chose to present it as a function of our contribution, discussed previously. Each point shows one algorithm on one of the datasets. There are three algorithms and six datasets, resulting in 18 points. It appears that there is a negative correlation between the time overhead of our method and its contribution, which can be explained as follows. Our method takes almost constant time regardless of the difficulty of the image pair. Epipolar geometry estimation algorithms, on the other hand run much faster on easy image pairs. Therefore, the larger our contribution, the easier it was for USAC/BEEM/BLOGS to finish running. Consider the extreme example of running USAC on the ``Open2 dataset''. The standard algorithm succeeded on only 27 image pairs, while with our preprocessing step 91 successes were registered (contribution of 239\%). This increase in performance was achieved at a factor of 2.7 in running times. As a result, due to the time overhead, we would recommend to apply our method only on challenging image pairs or when the standard method failed.

\subsection{Analysis}
\begin{figure}[htb]
    \centering
    	\includegraphics[width=0.45\textwidth]{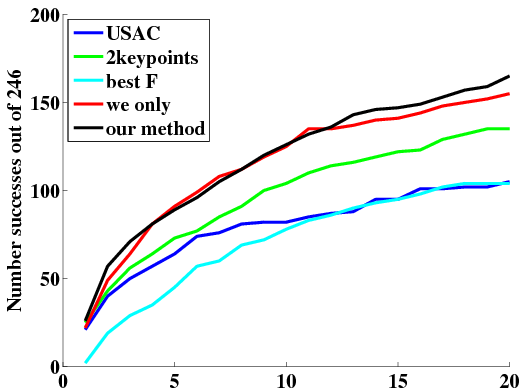}
    \caption{Analysis of different parts of our method. Major contribution can be attributed both to the 2keypoint matches generation and ranking and to the global ranking of matches.}
    \label{fig:Analysis}
\end{figure}

Our method is a combination of several steps. One could naturally ask whether all of these steps are necessary and what are their contributions. To answer those questions we tested several variations of the algorithm, which skip parts of our algorithm or replace them with standard ones. In Figure~\ref{fig:Analysis} we chose to present this analysis on the ``Open1 dataset'' while running USAC. Results on different datasets with other algorithms yielded qualitatively similar results.

The Blue curve is the result of the USAC algorithm without our contribution. The Green curve is the result when the 2keypoint match ranking is used as the input for epipolar geometry estimation. As a result we obtain better results than for the original USAC. This indicates that 2keypoint generation is a strong component of our method.

The Cyan curve is a result of altering our method. We generate $K_{2kp}(K_{2kp}-1)/2$ fundamental matrices, but instead of exploiting all of them, as it is done in our algorithm, we present here the fundamental matrix with the largest support as it is usually done. This shows really bad results, mostly because each fundamental matrix is calculated from four pairs of matches instead of 7 or 8, giving quite an inaccurate estimation. We believe that local optimization could yield better results, but this is beyond the scope of this work.

The Red curve is based on the putative matches $\left\{X\right\}$ and their $\left\{sfm\right\}$. Using the training set described above we converted the $sfm$ into a probability measure and used it as the input for epipolar geometry estimation. The resulting Red curve exhibits improved performance with respect to the 2keypoint match ranking. Therefore, this step also yields an important contribution to our method.

The Black curve shows the result of our entire method as is, without any changes. Those results are similar to those based on $\left\{X\right\}$ and their $\left\{sfm\right\}$. This is not surprising,  recalling the reasoning behind combining all the measures. As it was already mentioned, the intent was to improve performance on challenging image pairs, while not harming the performance on easy ones. Therefore, on challenging datasets such as the ``Open1 dataset'',  we expect our method to rely mostly on $\left\{X\right\}$ and their $\left\{sfm\right\}$ and have a minor contribution from $\left\{X_L\right\}$ and $\left\{X_B\right\}$. This is exactly what is exhibited by the similarity between the Black and Red curves. To verify the contribution of this step, we analyzed the performance of the algorithms on the easier ``BLOGS+USAC dataset''. We found that relying on putative matches $\left\{X\right\}$ alone, on average degraded the performance relative to our entire method for BLOGS/BEEM/USAC by 3.9\%, 3\% and 7.4\% respectively. Since combining all the measures is relatively cheap,  it is still recommended especially in the easy cases, even though its contribution is small.

Concluding this section we can state that every step of our algorithm is necessary and that our good results can mostly be attributed to the 2keypoint matches generation and ranking and to the global ranking of matches.

\section{Conclusions}\label{sect:conc}
In this paper we presented a general deterministic preprocessing step for epipolar geometry estimation algorithms.
It generates a set of putative feature matches between the two images, accompanied by prior probabilities that each match is correct.
The algorithm was tested on almost 900 image pairs from six publicly available datasets.
We showed experimentally that the results obtained by state-of-the-art algorithms which use the output of our algorithm outperform the same algorithms which uses the standard input. In general we can summarize that our preprocessing algorithm yields better results in hard cases and does not degrade on the easy ones.
This method is general and we believe that it can be used as the initial step of all guided RANSAC algorithms improving their performance.

%\begin{acknowledgements}
%\end{acknowledgements}

\bibliographystyle{spbasic}      % basic style, author-year citations
\bibliography{KS_bib}   % name your BibTeX data base
\end{document}